\documentclass[conference]{IEEEtran}

\usepackage{graphicx}
\usepackage{subcaption}
\usepackage{amsmath,amssymb,amsfonts,amsthm}
\usepackage{mathtools}
\usepackage{algorithmic}
\usepackage{algorithm}
\usepackage{xcolor}
\usepackage{tikzpagenodes}

\theoremstyle{definition}
\newtheorem{defn}{Definition}

\definecolor{darkgreen}{rgb}{0.0, 0.7, 0.0}

\usepackage[style=ieee, backend=biber]{biblatex}
\IEEEoverridecommandlockouts 

\begin{document}

\title{Autoencoder-Based Domain Learning for Semantic Communication with Conceptual Spaces}

\author{
\IEEEauthorblockN{Dylan Wheeler, Balasubramaniam Natarajan}
\IEEEauthorblockA{Kansas State University\\
Manhattan, Kansas, USA\\
Emails: dylan84@ksu.edu, bala@ksu.edu}
}

\maketitle

\begin{tikzpicture}[remember picture,overlay]
    \node[align=center] at ([yshift=1em]current page text area.north) {This work has been submitted to the IEEE for possible publication. Copyright may be\\ transferred without notice, after which this version may no longer be accessible.};
\end{tikzpicture}%

\vspace{-10pt}

\begin{abstract}
Communication with the goal of accurately conveying meaning, rather than accurately transmitting symbols, has become an area of growing interest. This paradigm, termed semantic communication, typically leverages modern developments in artificial intelligence and machine learning to improve the efficiency and robustness of communication systems. However, a standard model for capturing and quantifying the details of ``meaning'' is lacking, with many leading approaches to semantic communication adopting a black-box framework with little understanding of what exactly the model is learning. One solution is to utilize the conceptual spaces framework, which models meaning explicitly in a geometric manner. Though prior work studying semantic communication with conceptual spaces has shown promising results, these previous attempts involve hand-crafting a conceptual space model, severely limiting the scalability and practicality of the approach. In this work, we develop a framework for learning a domain of a conceptual space model using only the raw data with high-level property labels. In experiments using the MNIST and CelebA datasets, we show that the domains learned using the framework maintain semantic similarity relations and possess interpretable dimensions.
\end{abstract}


\IEEEpeerreviewmaketitle


\section{Introduction} \label{sec_introduction}

When we pause to take a look at the global state of connectivity and communication, we find ourselves in the midst of an ever-growing tidal wave of flowing information. As stated in the first page of the Ericsson Mobility Report \cite{2023_ericsson_mobilityReport}: ``As global mobile network data traffic continues to grow, with a compound annual growth rate (CAGR) of around 25 percent projected through 2028, smart network modernization becomes imperative.'' While this statement makes clear the challenges looming before us, it also hints at the common sentiment that the use of artificial intelligence (AI) and machine learning (ML) is a clear path forward toward addressing these challenges. These technologies have even begun to make their way into the mobile network standardization process \cite{2023_lin_artificialIn5GAdvanced3GPP}.

This sentiment has led to an explosion of interest in the notion of semantic communication. This paradigm shift leverages increased intelligence at various points in the communication chain to improve efficiency and performance. This is accomplished by focusing on the goal of accurately conveying meaning, rather than focusing solely on the goal of accurately transmitting symbols, as was initially suggested by Shannon and Weaver \cite{1949_shannonWeaver_mathTheoryOfComm}. For example, consider a speaker telling a listener, ``I drove a car down the road.'' In contrast, the speaker could have said ``I operated an automobile in such a way that it traversed along the path specifically designated for such a purpose.'' These two statements are essentially identical from a semantic view, though the first uses fewer words and is thus more efficient from a technical standpoint.\footnote{Note that though this example uses speech/text, this notion is modality-independent; similar semantically-equivalent-but-technically-distinct examples can be thought of for image-based communication, for instance.}

While the intuitive benefits of optimizing for semantic communication are readily seen, realizing these benefits from an engineering standpoint has proven to be a challenge. A plethora of different approaches have been put forth in the literature in recent years, with each having their own strengths and weaknesses \cite{2021_lan_semanticSurvey, 2022_iyer_semanticSurvey, 2023_wheeler_semComSurvey}. This stark diversity of ideas stems from a general lack of agreement on how to answer the question: how do we technically define and work with the concept of \textit{meaning}? To address this difficult topic, in our previous works \cite{2023_wheeler_semComLetter, 2023_wheeler_knowledgeDrivenSemcom}, we have proposed the use of \textit{conceptual spaces} (CSs), which are geometric-based models that can be used to organize and quantify meaning, originally put forth by cognitive scientist Peter G\"ardenfors \cite{2000_gardenfors_conceptSpace, 2014_gardenfors_semanticTheory}. Semantic communication with conceptual spaces (SCCS) can be both efficient and robust, while modeling meaning at an abstract level independent of the modality of communication.

One of the primary challenges of the proposed SCCS approach is obtaining the CS model. In our prior works, these models were hand-crafted using domain knowledge, after which a ML model used supervised learning to determine how to map raw data to the CS. This approach has two major drawbacks. First, the hand-crafted design of the CS is not scalable and depends on the knowledge of the designer. Second, ground-truth CS coordinate labels are required to train the semantic encoder, and obtaining this data can be difficult and costly. To address these shortcomings, in this work we propose a method of learning the domains of a CS in a partially supervised manner using a modified neural architecture based on the concept of an autoencoder. This approach removes the hand-crafted aspect and automates the CS design, requiring only high-level property labels for training, rather than fine-grained CS coordinate labels.


\subsection{Related Work} \label{subsec_relatedWork}

A multitude of ideas on how to approach the challenge of semantic communication have emerged over the last few years. Some take an information-theoretic approach and seek to extend Shannon's original work in a similar manner \cite{2023_ma_taskOrientedExplainableSemComm}. Others aim to capture meaning by way of knowledge graphs, which can then be used to facilitate semantic communication \cite{2023_xiao_reasoningOverAir}. However, the most popular approach by far has been to integrate deep learning (DL) technologies into the communication system, which can then learn semantic representations and effective communication strategies \cite{2023_qin_semCommMemory}. Although these early approaches have demonstrated success in improving communication in one way or another, they also have some shortcomings. DL-based systems lack interpretability, both in the operation of the components and the characteristics of the semantic representations. Moreover, there is no consensus on a framework for explicitly modeling and working with meaning, often leading to modality-specific systems and metrics \cite{2023_getu_makingSenseOfMeaningMetrics}. A common framework that is generally applicable across modalities and applications is required if we are to realize the practical benefits of semantic communication.

We advocate for such a framework in our previous studies \cite{2023_wheeler_semComLetter, 2023_wheeler_knowledgeDrivenSemcom}, where we introduce the use of the CS model for semantic communication. SCCS offers an interpretable, modality-independent method for modeling meaning in which semantic elements can be compared and differences in meaning can be quantified. In \cite{2023_wheeler_semComLetter}, we briefly introduce the framework and carry out some preliminary experiments, demonstrating the potential of the approach. In \cite{2023_wheeler_knowledgeDrivenSemcom}, we lay a solid theoretical foundation for SCCS and derive some upper bounds on the probability of a semantic error under the SCCS framework. We perform extensive experiments and compare our approach to the common DL-based end-to-end learning approach, demonstrating improved performance. As previously mentioned, however, both of these works involve hand-crafting the CS model, which is a major drawback with regards to ease of implementation and scalability.

Closely related to the problem of learning a CS model is the problem of \textit{feature learning} or \textit{representation learning}, where the goal is to extract meaningful patterns from data, usually in a lower-dimensional space. Some approaches to this problem include probabilistic models, manifold learning, and autoencoders \cite{2013_bengio_representationLearning}. As a simple example, the well-known principal component analysis (PCA) technique can be viewed as a particular example of all three of these methods \cite{2013_bengio_representationLearning}. However, these approaches are not directly applicable to the problem CS learning, as the learned dimensions often times lack interpretability and can be intractable.


\subsection{Contributions} \label{subsec_contributions}

To address the limitations outlined above, in this work we introduce an autoencoder-based technique for conceptual space (CS) learning. The key idea is to utilize semantic regularization based on the geometric structure of the learned representations and the semantic similarity derived from that structure to engender the key properties of a conceptual space in the learned space. Here, we focus on learning a single domain of a conceptual space and plan to build on this approach in future work. In short, our contributions include:
\begin{itemize}
    \item A novel approach to learning a domain of a CS model based on the autoencoder architecture. The modified architecture includes a maximum-similarity property classifier module to add semantic regularization.
    \item An iterative algorithm to learn the semantic properties and the model mapping raw data to the semantic domain.
    \item Experimental results using the MNIST and CelebA datasets to demonstrate that the proposed framework learns semantically consistent property regions and interpretable quality dimensions.
\end{itemize}


\section{Problem Definition} \label{sec_probDef}

In this work we focus on the problem of learning a single domain of a CS model, which we refer to as \textit{domain learning}. To formally define the problem, we first provide general definitions of CS components initially presented in \cite{2023_wheeler_knowledgeDrivenSemcom}.

\begin{defn}[Quality Dimension] \label{def_QualDim}
    A quality dimension is a set of scalar values quantifying some quality of an idea or object. Denoted by $\mathcal{Q}$, specific values along a quality dimension are denoted by $q$, such that $q \in \mathcal{Q}$.
\end{defn}

\begin{defn}[Domain] \label{def_Domain}
    A domain $\mathcal{D}$ is a set that is constructed from the Cartesian product of integral quality dimensions, i.e., $\mathcal{D} = \bigtimes_{n=1}^N Q_n$.  A point within a domain is specified by a column vector of quality values $\textbf{q} = \begin{pmatrix} q_1 & q_2 & \cdots & q_N \end{pmatrix}'$. 
\end{defn}

\begin{defn}[Property] \label{def_Property}
    A property $\mathcal{P}$ of a single domain $\mathcal{D}$ is defined as a convex subset of that domain, i.e., $\mathcal{P} \subset \mathcal{D}$ where, for $\lambda \in [0,1]$ and any $\textbf{p}_1, \textbf{p}_2 \in \mathcal{P}$, $\lambda \textbf{p}_1 + (1-\lambda) \textbf{p}_2 \in \mathcal{P}$.
\end{defn}

Quality dimensions are the building blocks of CS models. Integral quality dimensions form a domain, where two qualities are said to be \textit{integral} if a value on one dimension cannot be assigned to an object without giving it a value on the other, e.g., hue and brightness of color \cite{2000_gardenfors_conceptSpace}. Properties are then defined as regions within a domain, where the convexity requirement ensures that if two points represent a given property, all points in between also represent that property.

As a geometric entity, it is natural to think of a semantic distance or distortion metric on a domain. Here we consider a particular case of the general semantic distortion presented in \cite{2023_wheeler_knowledgeDrivenSemcom} with only a single domain. 

\begin{defn}[Semantic Distortion]
    Semantic distortion is a function $\delta$ mapping two points in the domain $\mathcal{D}$ to the non-negative real line, i.e., $\delta: \mathcal{D} \times \mathcal{D} \rightarrow \mathbb{R}_+$.
\end{defn}

Moreover, we define the notion of semantic similarity, which acts on the semantic distortion to yield a measure of similarity between two points in the CS. In \cite{2000_gardenfors_conceptSpace}, it is suggested that a Gaussian function could be used to transform semantic distance to similarity; we adopt this function here.

\begin{defn}[Semantic Similarity]
    Semantic similarity maps semantic distortion to the interval (0,1], with 0 representing total dissimilarity and 1 representing identity. Formally, we have $\sigma : \mathbb{R}_+ \rightarrow (0,1]$ such that,
    \begin{equation}
        \sigma(d_{ij}) = e^{-cd_{ij}^2},
    \end{equation}
    where $d_{ij}$ is the distortion between two points $\textbf{q}_i$ and $\textbf{q}_j$ and $c$ is a tunable parameter.
\end{defn}

\begin{figure}[t]
    \centering
    \includegraphics[width = 0.37\textwidth]{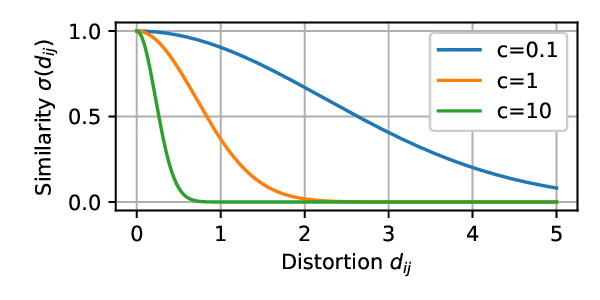}
    \caption{Semantic similarity vs. semantic distortion.}
    \label{fig_similarities}
\end{figure}

We see that the similarity is 1 when $d_{ij}$ is 0 and approaches 0 as $d_{ij}$ approaches infinity. Similarity will remain approximately 1 for small distortion values, begin to quickly decrease as distortion continues to increase, and eventually become approximately 0 for large distortion values. The parameter $c$ can be used to ``shrink'' or ``stretch'' this function, such that the similarity function is more or less sensitive to small changes in distortion, as shown in Figure \ref{fig_similarities}.

At the outset, we assume that we are given a dataset with two components. First, the dataset contains samples of the raw data to be communicated, i.e., the input to the semantic encoder in a typical semantic communication setup. Second, it contains \textit{semantic property} labels for each data sample. Note that these labels contain no information about the property regions as defined in Definition \ref{def_Property}, and only specify \textit{which} property the data possesses. Further, in this work we assume that the quality dimensions to be learned are continuous intervals that lie in Euclidean space such that $\mathcal{Q} \subset \mathbb{R}$ and $\mathcal{D} \subset \mathbb{R}^N$ where $N$ is the number of dimensions in the domain. As such, we take the semantic distortion function to be the Euclidean distance function, i.e.,
\begin{equation}
    \delta(\textbf{q}_i,\textbf{q}_j) = \Vert \textbf{q}_i - \textbf{q}_j \Vert_2 = \sqrt{ \sum_{n=1}^N \left( q_i^{(n)} - q_j^{(n)} \right)^2 }.
\end{equation}

Given these data and assumptions, our objective is to employ ML to learn a CS domain describing the semantics of the data. Thus, the quality dimensions will be characterized by the dimensions that the network learns to map the data to. To validate the quality dimensions of the CS domain, we require them to be interpretable. Furthermore, we require that semantically similar properties be located closer in the space than semantically dissimilar properties. Instead of learning the property regions themselves, we can take these regions to be the corresponding Voronoi regions formed using the semantic distortion function and a set of $P$ property \textit{prototypes} $\mathcal{P} = \{\textbf{p}_i\}_{i=1}^P$, where $\textbf{p} \in \mathcal{D}$ \cite{2000_gardenfors_conceptSpace, 2023_wheeler_knowledgeDrivenSemcom}. Thus, specifying the property regions is equivalent to specifying the coordinates of each prototype point in the domain, and all we need to learn are these prototype points. To summarize, we have three requirements for the result of the domain learning process:
\begin{enumerate}
    \item[R1.] A model mapping data to interpretable dimensions.
    \item[R2.] A set of distinct points defining the property regions.
    \item[R3.] Semantically consistent similarities between properties.
\end{enumerate}


\section{Proposed Approach} \label{sec_propApproach}

\begin{figure}
    \centering
    \includegraphics[width=0.41\textwidth]{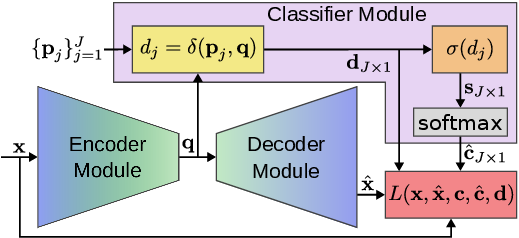}
    \caption{Architecture of the domain learning network.}
    \label{fig_domainLearningArch}
\end{figure}

To address these challenges, we propose an autoencoder-based approach as shown in Figure \ref{fig_domainLearningArch}. The architecture contains three primary modules: the encoder, the decoder, and the classifier. The encoder and decoder modules are the same as in a traditional autoencoder. The central innovation is the addition of the property classifier module, which provides a kind of \textit{semantic regularization} to the network while learning. The details of each module are outlined below.

\subsection{Encoder Module}
\label{subsec_encoder}

The encoder takes the input and reduces the dimensionality to produce an output vector $\textbf{q}$, which is a lower-dimensional representation of the input data $\textbf{x}$. This module consists of traditional neural network layers, which can vary depending on the input data type. For example, the encoder can comprise many dense layers for typical vector-shaped input data, or convolution layers for three-dimensional image data. At the output of the encoder, a dense layer with no activation is included to reduce the features to the domain dimensionality.


\subsection{Decoder Module}
\label{subsec_decoder}

The task of the decoder module is to reconstruct the input data using the output of the encoder module. In the domain learning task, the decoder module is responsible for ensuring that similar data samples are mapped closer together within the CS domain. The intuition is that if two inputs $\textbf{x}$, $\textbf{x}'$ are similar, the decoder will require similar representations $\textbf{q}, \textbf{q}'$ in order to recover the original inputs $\textbf{x}$ and $\textbf{x}'$. Structurally, the decoder module is built to mirror the encoder module, with layers that expand the dimensions reduced by layers of the encoder. To measure the quality of the reconstructed output $\hat{\textbf{x}}$, we define the reconstruction loss term $\ell_r$ as the mean squared difference between the input and reconstructed data,
\begin{equation} \label{eq_reconstruction_loss}
    \ell_\text{r}(\textbf{x},\hat{\textbf{x}}) = \frac{1}{M}\sum_{m=1}^M (x_m - \hat{x}_m)^2.
\end{equation}

\begin{figure*}
    \centering
    \includegraphics[width=0.75\textwidth]{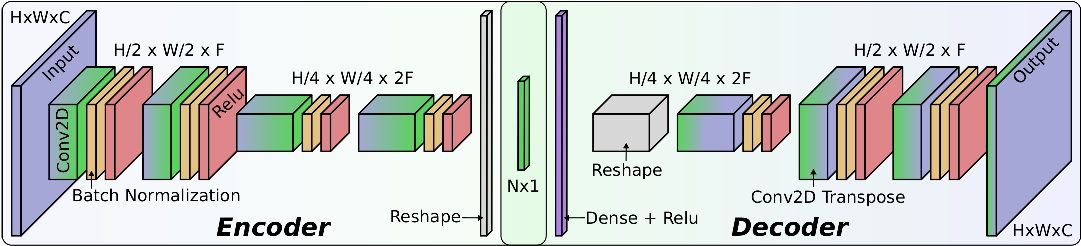}
    \caption{Architecture of the encoder and decoder used in the experiments. \vspace{-.4cm}}
    \label{fig_encoder_decoder}
\end{figure*}


\subsection{Classifier Module}
\label{subsec_classifier}

The classifier module is the key element of the domain learning architecture. This module takes as inputs the encoded representation $\textbf{q}$ and a list of prototypes $\{\textbf{p}_j\}_{j=1}^J$ that describe the prototypical properties within the domain. First, the semantic distortion between $\textbf{q}$ and each prototype is computed, yielding a $J\times1$ vector $\textbf{d}$ of distortion values. The semantic similarity is then computed for each of the distortion values, the result of which is another $J\times1$ vector $\textbf{s}$. Finally, a softmax operation is performed on $\textbf{s}$ to get a classification vector $\hat{\textbf{c}}$, which is passed to the loss function. Note that the classifier module does not contain any trainable layers.

The classifier module acts as a \textit{maximum-similarity decoder} mapping a given domain representation to the property within the domain for which it is the most similar. As the properties basically encode the difference in meanings between the input data samples, this module ensures that the encoded representations are separated in the space according to their different meanings. Furthermore, unlike a basic classifier network, the output classification is entirely dependent on the similarity between the property prototypes and the domain representation. This ensures that notions of semantic distortion and semantic similarity are well-represented within the learned domain.

There are two loss terms associated with the classifier module. First, the classification loss seeks to train the network to produce classification vectors $\hat{\textbf{c}}$ which are close to the true $\textbf{c}$ representing the correct properties, which are one-hot encoded vectors. Categorical cross-entropy is used for this loss term,
\begin{equation} \label{eq_classification_loss}
    \ell_c (\textbf{c}, \hat{\textbf{c}}) = -\sum_{j=1}^J c_j \log(\hat{c}_j).
\end{equation}

Second, we need to ensure that the distances between representations in the space are reasonable, i.e., the similarities take on meaningful values. For example, in Figure 1, for the $c=1$ curve, any distortions greater than approximately 2.25 will yield a similarity of essentially zero. Thus, we include a distortion-regularization term which penalizes large distortions between an encoded representation and the prototypes of the properties to which it does \textit{not} belong. This ensures that the space does not ``blow up'' and that the regions remain relatively close together. We achieve this with the function
\begin{equation} \label{eq_distance_reg}
    \ell_d(\textbf{d}) = \frac{1}{J-1} \sum_{j\neq j^*} \frac{1}{k_1} d_j^{k_2},
\end{equation}
where $j^*$ denotes the index of the true property, and $k_1$ and $k_2$ are hyperparameters that can be used to adjust the range of distortion values that are penalized.


\subsection{Putting it Together: Domain Learning Algorithm}

Combining equations (\ref{eq_reconstruction_loss})-(\ref{eq_distance_reg}), the loss function used to train the domain learning architecture is defined as
\begin{equation}
    L(\textbf{x}, \hat{\textbf{x}}, \textbf{c}, \hat{\textbf{c}}, \textbf{d}) = \alpha \ell_r(\textbf{x},\hat{\textbf{x}}) + \beta \ell_c(\textbf{c},\hat{\textbf{c}}) + \lambda \ell_d(\textbf{d})
\end{equation}
where $\alpha, \beta$ and $\lambda$ are hyperparameters that can be tuned to balance the terms of the loss function. This loss is used with backpropagation to tune the parameters of the encoder and decoder while learning the domain. However, recall that computing $\textbf{d}$ requires the set of prototypes $\{\textbf{p}_j\}_{j=1}^J$, which are one of the aspects of the domain which we are trying to learn and do not have \textit{a priori} knowledge of at the start. Therefore, we propose an iterative algorithm to learn both the dimensions of the domain and the property prototypes in a joint manner; this algorithm is presented in Algorithm \ref{alg_main}.

\begin{algorithm} 
\caption{Domain Learning}
\begin{algorithmic}[1] \label{alg_main} 
    \REQUIRE $\alpha, \beta, \lambda, \mu, M, B, J, \{\textbf{p}_j\}_{j=1}^J = \textbf{0}$
    \FOR{each epoch $i$ in $\{1,2,\ldots,E\}$}
        \FOR{each batch $m$ in $M$ random batches}
            \IF{$i$ is 1}
                \STATE $L \leftarrow \frac{1}{B} \sum_{b=1}^B \alpha\ell_r(\textbf{x}_b,\hat{\textbf{x}}_b) + \lambda \ell_d(\textbf{d}_b)$
            \ELSE
                \STATE $L \leftarrow \frac{1}{B} \sum_{b=1}^B \alpha\ell_r(\textbf{x}_b,\hat{\textbf{x}}_b) + \beta \ell_c(\textbf{c}_b,\hat{\textbf{c}}_b) + \lambda \ell_d(\textbf{d}_b)$
            \ENDIF
            \STATE Perform batch update with backpropagation
        \ENDFOR
        \STATE Encode $MB$ random samples to obtain $\{\textbf{q}\}$
        \FOR{each property $j$ in \{1,2,\ldots,J\}}
            \STATE $\overline{\textbf{q}}_j \leftarrow \text{mean}(\{\textbf{q} : \textbf{q} \text{ possesses property } j\})$
            \STATE $\textbf{p}_j^{(i)} \leftarrow \overline{\textbf{q}}_j $ if $i = 1$, else $\mu\textbf{p}_j^{(i-1)}+ (1-\mu)\overline{\textbf{q}}_j $
        \ENDFOR
    \ENDFOR
    \RETURN Encoder, decoder and prototypes $\{\textbf{p}_j\}_{j=1}^J$
\end{algorithmic}
\end{algorithm}

\begin{figure*}[h]
    \centering
    \begin{subfigure}[b]{0.32\textwidth}
        \includegraphics[width=\linewidth]{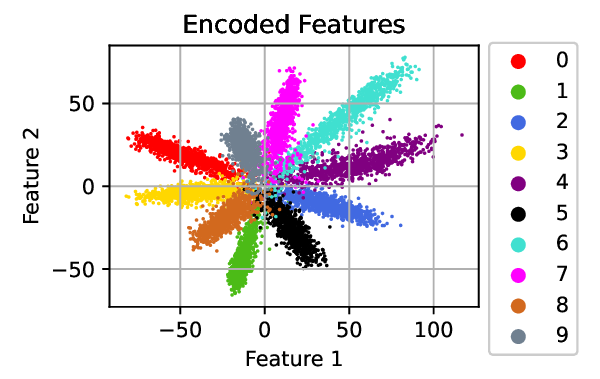}
        \caption{Basic classifier network.}
        \label{subfig_classifier_features}
    \end{subfigure}
    \hfill
    \begin{subfigure}[b]{0.32\textwidth}
        \includegraphics[width=\linewidth]{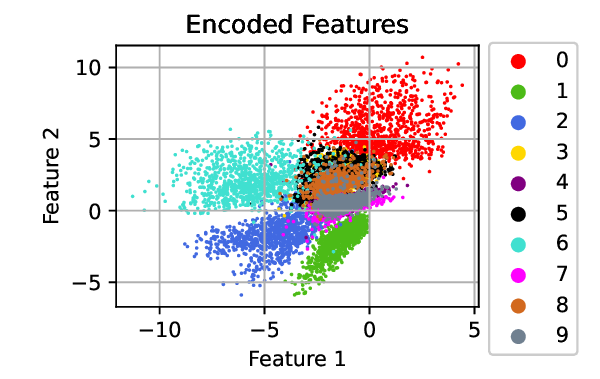}
        \caption{Basic autoencoder network.}
        \label{subfig_autoencoder_features}
    \end{subfigure}
    \hfill
    \begin{subfigure}[b]{0.32\textwidth}
        \includegraphics[width=\linewidth]{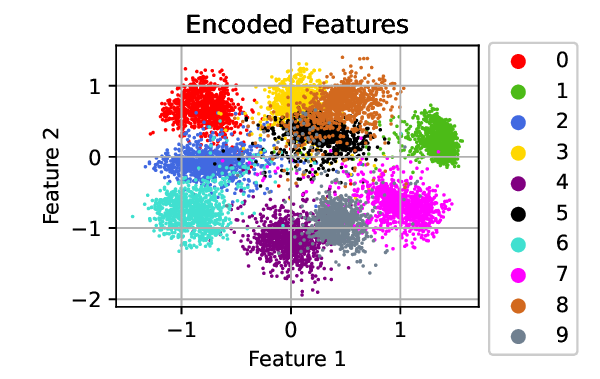}
        \caption{Proposed domain learning approach.}
        \label{subfig_learner_features}
    \end{subfigure}
    \caption{The learned representations of the validation samples of the MNIST dataset for (a) a basic classifier (b) a basic autoencoder and (c) the proposed domain learning framework. \vspace{-.4cm}}
    \label{fig_mnist_features}
\end{figure*}

First, the number of batches $M$ and batch size $B$ used for training on each step of the algorithm is set. Then the encoder and decoder are trained for one step with the parameter $\beta$ set to 0; this is equivalent to training a vanilla autoencoder on $MB$ samples of the input data with some distortion regularization. For this first step, prototypes are set to the zero vector $\textbf{0}$ to keep the autoencoder-learned features centered at the origin. To initialize the prototype points, $M$ batches are then passed through the encoder to obtain $MB$ encoded representations. The representations that correspond to property $j$ are then averaged to obtain the initial prototype point for that property. With the initial properties set, the network is then trained over another $M$ batches using the full loss function, after which the prototypes are updated in the same manner that they were initialized, with the one difference that a parameter $\mu \in [0,1]$ is used to ``mix'' the old prototypes with the new ones. This process repeats for a fixed number of epochs, denoted by $E$.


\section{Experimental Results} \label{sec_expResults}

We use image data in our experiments, as images contain semantics that are easy to identify and interpret and have well-established architectures for encoding and decoding, such as convolutional neural networks (CNNs). Note that the proposed approach is not exclusive to images, and other modalities can be used with an appropriate model architecture. 

The architecture of the encoder and decoder is shown in Figure \ref{fig_encoder_decoder}. The encoder uses four convolutional layers, each followed by batch normalization and relu activation. The first convolutional layer has $F$ filters, a kernel size of $7\times7$ and a stride of $2$, halving the input dimensions. The second layer has $F$ filters with a $5\times5$ kernel and stride of one. The third and fourth layers both use $2F$ filters with $3\times3$ kernels, while the third layer uses a stride length of $2$. The resulting tensor is reshaped to a single vector, which is then passed through a dense layer with no activation to obtain the encoded domain representation. The decoder is a mirror image of the encoder, using transposed convolutional layers to ``undo'' the convolutional layers of the input. Experiments are performed using Tensorflow, and the Adam optimizer with a learning rate of 0.0002 is used to train the networks. For each experiment, the batch size $B$ is set to 32, the number of batches $M$ used to train each epoch is 250, and the number of epochs is 75.


\subsection{Experiment 1: MNIST} \label{subsec_mnist}

We use the MNIST dataset as a starting point. The MNIST dataset contains 70,000 28x28x1 black-and-white images of handwritten digits from 0-9, split into 60,000 training images and 10,000 test images \cite{2012_deng_MNIST}. Thus, we have $H=28$, $W=28$, and $C=1$. As the MNIST dataset is relatively simple, we choose $F=16$ for encoder/decoder filter parameter. For the loss function hyperparameters, we choose $\alpha=5$, $\beta = 1$ and $\lambda = 1$ so that the three loss terms take on a similar range of values during training. The prototype mixing parameter is chosen to be $\mu = 0.75$ to encourage smooth updates. For the scaling parameters in equation $\ref{eq_distance_reg}$, we use $k_1 = 50$ and $k_2 = 5$.

We set the dimension of the encoded representations to $N=2$. In addition to the domain learning model, we also train a basic classifier (encoder only, with a single dense classifier head layer) and a basic autoencoder (encoder and decoder only). Figure \ref{fig_mnist_features} shows the learned feature spaces for the three approaches. In Figure \ref{subfig_classifier_features}, we see that the classifier learns to map the different digits to distinct regions, but the semantic similarity is not captured, e.g., 4 and 9 are very similar, but their regions are separated by 6 which is not similar to either. For the autoencoder features shown in Figure \ref{subfig_autoencoder_features}, the similarity is captured, but the properties are not distinct, e.g., 4, 7, and 9 are all mapped to approximately the same region of the space. In Figure \ref{subfig_learner_features}, we see that similar properties are indeed mapped closer to each other (e.g., 4 and 9, and 3 and 5) and that each property has a distinct region within the space.

These results show that the proposed approach is able to satisfy R2 and R3. However, the axes are not very interpretable. This is due to the nature of the input data; what numerically quantifiable quality makes a 4 similar to a 9 but different than a 3? What is ``in between'' a 4 and a 6? In order to learn continuous dimensions about which the semantics of the data vary, the data must share such a continuous quality.


\subsection{Experiment 2: CelebA} \label{subsec_celeba}

\begin{figure}
    \centering
    \includegraphics[width=0.85\linewidth]{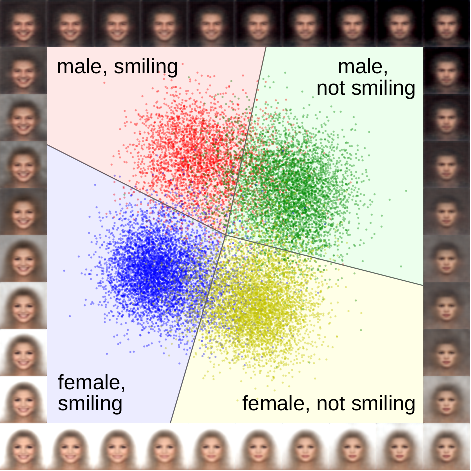}
    \caption{The learned representations of the validation samples of the CelebA dataset, and decoded representations at the edge of the domain. The domain encodes smiling/not smiling primarily along the x-axis and female/male primarily along the y-axis. \vspace{-.2cm}}
    \label{fig_celeba_results}
\end{figure}

Next, we utilize the CelebA dataset for domain learning. The dataset contains approximately 200,000 color images of celebrity faces \cite{2015_liu_celeba}, which we resize to 112x112x3. The labels contain 40 binary attribute annotations, such as male/female or ``has blonde hair''/``does not have blonde hair''. We select two attributes to form our property labels, namely, male/female and smiling/not smiling, due to the attributes being evenly represented (in contrast to ``has gray hair'', which has far more negative samples than positive). In this experiment, we increase the filter parameter to $F = 32$, set $\alpha=1, \beta=2, \lambda=1$, and $\mu=0.75$, while $k_1$ and $k_2$ are the same.

Figure \ref{fig_celeba_results} shows the learned features of the ``face'' domain, where we again choose $N=2$. We see that as in the case of MNIST, the model learns property regions that are both distinct and accurately characterize the semantic similarity between properties. For example, the center of the ``female, not smiling'' property region is closer to the centers of the ``male, not smiling'' and ``female, smiling'' regions than that of the ``male, smiling'' region. This again demonstrates how the proposed framework satisfies R2 and R3.

Further, the learned dimensions of the domain are indeed interpretable, satisfying R1. It holds that moving from left to right on the x-axis corresponds to decreasing the amount of smiling, regardless of male or female. Similarly, moving from the bottom up on the y-axis corresponds to transitioning from female to male faces, regardless of whether the face is smiling. This is seen on the border of Figure 5, which shows decoded representations at the extreme points on edge of the domain. These images are generated directly from the selected domain coordinates, i.e., no real images were used to create them. This illustrates a powerful feature of the proposed autoencoder-based framework, namely, that the decoder module can be used to \textit{interpret} the learned dimensions.

Finally, note that the encoded representations require far less bits as opposed to the raw images themselves. As only two floating-point values are used to encode the semantics of the images, only $2\times32=64$ bits are required to convey the relevant meaning, rather than the $112\times112\times3\times8=301,056$ bits (before compression) required to transmit the raw image. This corresponds to a rate reduction of over 99\%, illustrating the gains that are possible using semantic communication.


\section{Conclusion} \label{sec_conclusion}

In this work, we propose a method for learning a domain of a CS model, to reduce the reliance of SCCS on hand-crafted semantic models. The proposed framework is based on the autoencoder architecture, with an added property classifier module to provide semantic regularization during training. We set out to show that the proposed method learns distinct property regions that preserve semantic similarity relations, and that the learned dimensions of the domain are interpretable. Experimental results confirmed that the framework is indeed capable of meeting these requirements of CS learning.

There are multiple directions of future research that can build off of this work. First, the most obvious step forward involves expanding to the learning of a complete CS model with multiple domains. Second, only linear dimensions and Euclidean semantic distortion were considered here, while CS models can be more general. How to handle general dimensions and distances is an interesting challenge. Finally, the framework is generally not limited to images, which were chosen for their ease of interpretability. It would be interesting to see how the framework behaves given other modalities to learn on, such as text and audio.

\printbibliography

\end{document}